\documentclass[sigconf]{acmart}

\usepackage{booktabs}
\usepackage{multirow}
\usepackage{array}
\usepackage{xcolor}
\usepackage{microtype}
\usepackage{amsmath}

\definecolor{grpobg}{RGB}{240,253,244}
\definecolor{failbg}{RGB}{255,241,242}

\begin{document}

\title{Ontology Concept Overlap as a Training Signal:\\
Knowledge-Grounded Reinforcement Learning for Clinical Question Answering}

\author{Aditya Tanna}
\authornote{Corresponding author}
\email{202103023@dau.ac.in}
\affiliation{%
  \institution{Dhirubhai Ambani University}
  \city{Gandhinagar}
  \state{Gujarat}
  \country{India}
}

\author{Abhishek Jindal}
\email{abhishek\_jindal@dau.ac.in}
\affiliation{%
  \institution{Dhirubhai Ambani University}
  \city{Gandhinagar}
  \state{Gujarat}
  \country{India}
}

\copyrightyear{2026}
\acmYear{2026}
\setcopyright{cc}
\setcctype{by}
\acmConference[CIKM '26]{Proceedings of the 35th ACM International Conference on Information and Knowledge Management}{November 07--11, 2026}{Rome, Italy}
\acmBooktitle{Proceedings of the 35th ACM International Conference on Information and Knowledge Management (CIKM '26), November 07--11, 2026, Rome, Italy}
\acmDOI{10.1145/3799682.3840057}
\acmISBN{979-8-4007-2539-5/2026/11}

\begin{abstract}
Reinforcement learning post-training for language models relies on two reward designs: human preferences (RLHF, DPO) and binary verifiers (RLVR). Clinical question answering fits neither. Near-correct answers differ by a single substituted entity, and no executable check decides clinical correctness. We instantiate a soft verifier from a maintained controlled vocabulary: UMLS Concept Unique Identifier overlap (via scispaCy, set-level F1) gives a graded, externally specified reward computed without a model in the loop. We combine it inside GRPO with an entropy-normalised LLM judge, which covers the safety and evidence axes overlap cannot see, and a small consistency penalty on padding and repetition that keeps early-training samples scorable. This three-term composite improves over SFT on Phi-3-mini (3.8B) over MedQA by 2.9\% on EM (0.700 vs 0.680) and 39\% on Token-F1 (0.202 vs 0.145); on Llama-3.2-3B the corresponding gains are 14\% on EM and 35\% on Token-F1. We report Token-F1 as the primary metric because it credits partially-correct clinical content that EM discards at this open-generation scale. Main-table results are means over 3 seeds with standard deviations below 0.005. The method transfers to PubMedQA, where training on the PubMedQA train set with the same composite reward improves Token-F1 over SFT by 22\% on Phi-3-mini and 17\% on Llama-3.2-3B without retuning. A reward ablation on Phi-3, varying the judge--ontology split at a fixed consistency weight, attributes 3 EM points to the ontology term, the contribution that catches entity substitutions the judge cannot. Three negative findings constrain the design: DPO under random negatives underperforms SFT for strong-prior models but helps the weakest-prior one; PPO under a sparse neural reward diverges; GRPO with KL-in-loss collapses at 7B. The recipe is designed to transfer to any domain with a maintained controlled vocabulary, a generalisation we leave to future work.
\end{abstract}



\maketitle

\section{Introduction}

Naming the right drug class but the wrong specific compound is a clinical error. Cross-entropy loss does not see the difference. It assigns the same penalty to ``consider an anticoagulant'' as to ``warfarin'' when the answer is ``heparin.''

Reinforcement learning can reward correctness directly, but only when the reward signal measures correctness. The dominant designs do not fit clinical question answering. RLHF and DPO~\cite{rafailov2023dpo} rely on human-preference pairs; near-correct and correct clinical responses differ by a single substituted entity, and random-negative pairs do not capture that contrast~\cite{azar2024}. RLVR~\cite{lambert2024tulu3,guo2025deepseekr1} replaces the reward model with a verifier, but no executable check decides whether ``ceftriaxone'' is the right empirical antibiotic for meningitis. Recent ontology-aware approaches~\cite{khatwan2025kgreward} use UMLS knowledge graph paths as rewards for diagnostic reasoning, but require graph traversal that is brittle on noisy generations.

We propose to instantiate the verifier from a maintained ontology, scored at the concept level rather than via graph traversal. Where RLVR uses an executable check that returns a binary pass, ours is a soft verifier: it returns graded overlap against an external vocabulary, which suits a domain where correctness is a matter of degree and no exact check exists. The Unified Medical Language System (UMLS)~\cite{bodenreider2004umls} assigns Concept Unique Identifiers (CUIs) that collapse synonymous strings (``heart attack,'' ``myocardial infarction,'' ``MI'') onto one identifier and keep clinically distinct entities (``heparin,'' ``warfarin'') apart. CUI overlap between a generated and a reference response gives a graded, machine-checkable, externally specified reward. We compute it via scispaCy~\cite{neumann2019scispacy}, pair it with an entropy-normalised LLM judge that covers safety and evidence axes overlap alone cannot capture, and run the composite reward inside GRPO~\cite{shao2024deepseekmath}.

On Phi-3-mini (3.8B) over MedQA~\cite{jin2021medqa}, the composite reward improves over SFT by 2.9\% on EM (0.700 vs 0.680) and 39\% on Token-F1 (0.202 vs 0.145); on Llama-3.2-3B the gains are 14\% EM and 35\% Token-F1, with main-table means computed over three seeds and standard deviations under 0.005. We treat Token-F1 as the primary metric: on open-ended clinical generation, EM discards a response that names the right entity inside a longer answer, whereas Token-F1 credits the overlapping clinical content (\S\ref{sec:experiments}). The method transfers to PubMedQA~\cite{jin2019pubmedqa}: under the same composite reward, GRPO improves Token-F1 over SFT by 22\% on Phi-3-mini and 17\% on Llama-3.2-3B without retuning the reward mixture. A reward ablation on Phi-3 attributes 3 EM points to the ontology term specifically: it catches entity substitutions (``penicillin G'' for ``ceftriaxone'') that the judge cannot distinguish from correct responses on fluency and evidence axes. Three negative findings constrain the design space.

\textbf{Contributions.} We instantiate a soft verifier from a curated ontology as a reward source for language-model RL, distinct from preference rewards, from binary RLVR checks, and from KG-path rewards~\cite{khatwan2025kgreward} that require graph traversal; we run UMLS CUI overlap inside GRPO with an ablation that isolates the ontology component as responsible for entity-level corrections; we validate across model families and seeds and transfer to a second QA benchmark.

\section{Method}
\label{sec:method}

\subsection{Composite reward}

For each (question $q$, generated response $y$, reference $y^*$) triple, the composite reward combines three terms:
\begin{equation}
R(q, y, y^*) = \alpha\,R_\text{judge}(q, y) + \beta\,R_\text{CUI}(y, y^*) + \gamma\,R_\text{cons}(y),
\label{eq:composite}
\end{equation}
with $\alpha=0.6$, $\beta=0.3$, $\gamma=0.1$. The judge--CUI split between $\alpha$ and $\beta$ is set by Spearman correlation against held-out EM (Appendix~\ref{app:calibration}); the small consistency weight $\gamma$ is fixed rather than tuned. We found this mixture trained more stably than the judge--CUI pair alone: the consistency term suppresses the padding and repetition that the policy otherwise learns to exploit early in training, before the CUI signal has enough coherent generations to score.

\textbf{Ontology component $R_\text{CUI}$.} scispaCy's EntityLinker extracts UMLS CUIs from both $y$ and $y^*$. With $\mathcal{C}_\text{gen}$ and $\mathcal{C}_\text{gold}$ the two CUI sets, the reward is set-level F1:
\begin{equation}
R_\text{CUI} = \frac{2|\mathcal{C}_\text{gen}\cap\mathcal{C}_\text{gold}|}{|\mathcal{C}_\text{gen}|+|\mathcal{C}_\text{gold}|}.
\label{eq:cui}
\end{equation}
Synonymous expressions share a CUI; distinct clinical entities do not. The signal is graded in $[0,1]$ and computed without any model in the loop. Because the reward depends on naming the right entity rather than on surface fluency, we hypothesise it is harder to game by hedging or padding than a learned reward model; we leave an adversarial reward-hacking probe to future work. scispaCy's linker has precision--recall trade-offs (Appendix~\ref{app:calibration}); the calibration analysis shows the resulting noise is small enough that the gradient signal stays informative.

\textbf{LLM-judge component $R_\text{judge}$.} A frozen Phi-3-mini scores each response on four axes (clinical accuracy, terminology, evidence, safety) and returns a probability distribution over $\{0,\ldots,5\}$. We reward distributions concentrated on high scores via normalised entropy:
\begin{equation}
R_\text{judge} = 1 - H(p)/\log_2 6, \quad H(p) = -\textstyle\sum_k p_k \log_2 p_k.
\end{equation}
The judge covers safety and evidence axes that entity overlap cannot detect; its known gameability~\cite{singhal2024lengthrlhf} means the ontology component must complement it rather than replace it.

\textbf{Consistency component $R_\text{cons}$.} A lightweight penalty on degenerate generations: it scores down responses that pad or repeat, computed from the rate of repeated $n$-grams in $y$ and capped to $[0,1]$. It uses no model and no reference. The term carries a small weight ($\gamma=0.1$) and does not encode clinical correctness; its job is to keep early-training samples coherent enough for the judge and CUI terms to score meaningfully, which is where the stability gain comes from.

\subsection{GRPO with composite reward}

\begin{figure*}[t]
\centering
\includegraphics[width=0.66\textwidth]{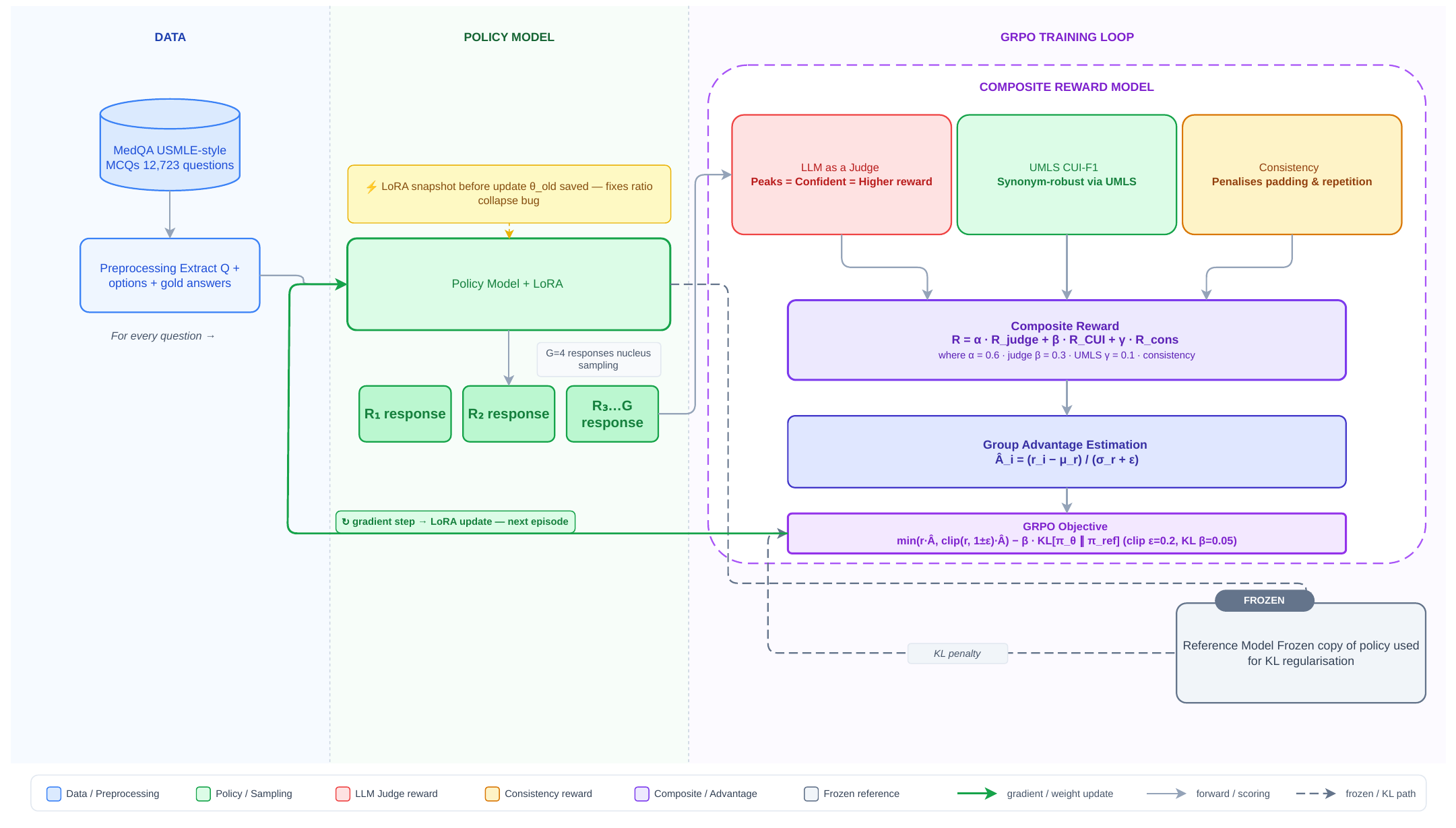}
\caption{Composite-reward GRPO. The policy samples $G{=}4$ responses per question; each is scored by the frozen LLM judge, by UMLS CUI overlap against the reference, and by a consistency penalty on padding and repetition. The three terms are combined as in Eq.~\ref{eq:composite} ($\alpha{=}0.6$, $\beta{=}0.3$, $\gamma{=}0.1$). Group-normalised advantages update the LoRA adapter. Reference and judge remain frozen throughout training.}
\label{fig:pipeline}
\end{figure*}

We use Group Relative Policy Optimization~\cite{shao2024deepseekmath} because it accepts sequence-level rewards without a value network. For prompt $q$ with $G$ sampled responses, group-normalised advantages are $\hat{A}_i = (r_i - \mu_r)/(\sigma_r + \varepsilon)$ with $\mu_r$ the within-group mean. The clipped surrogate follows the standard form with $\beta_\text{KL}=0.05$, clip $\epsilon=0.2$, $G=4$ samples at $T=0.7$. PPO~\cite{ouyang2022instructgpt}, by contrast, needs a value network to decompose one sequence-level reward into per-token advantages without per-step signal; \S\ref{sec:negative} shows this fails for our reward, whereas GRPO's group normalisation keeps single-GPU training with episode-level rewards tractable. Figure~\ref{fig:pipeline} summarises the pipeline.

\section{Experiments}
\label{sec:experiments}

\subsection{Setup}

\textbf{Datasets.} MedQA (USMLE 4-options)~\cite{jin2021medqa}, 12{,}723 questions (10{,}178 train / 1{,}272 dev / 1{,}273 test). PubMedQA~\cite{jin2019pubmedqa} for transfer evaluation. The 500 zero-shot outputs used for reward calibration (Appendix~\ref{app:calibration}) are drawn from the training split and are disjoint from the 500 held-out test questions used in all reported results; $\alpha$ is never tuned on the evaluation set.

\textbf{Models.} Three families with LoRA ($r{=}8$, $\alpha{=}32$): Phi-3-mini-4k~\cite{abdin2024phi3} (3.8B), Qwen2.5-7B~\cite{qwen25} (7B), and Llama-3.2-3B~\cite{llama3} (3B).

\textbf{Training.} Single A100 80GB, 2{,}000 steps, AdamW. GRPO LR $10^{-5}$ (3B/3.8B), $5{\times}10^{-7}$ (7B). DPO $\beta{=}0.1$, LR $10^{-5}$. SFT LR $2{\times}10^{-5}$ (3B/3.8B), $5{\times}10^{-6}$ (7B). Three seeds per (model, method): $\{42, 123, 456\}$ control LoRA initialisation and the order of training examples; hyperparameters are fixed across seeds. Results are reported as mean and standard deviation.

\textbf{DPO baseline construction.} We sample rejected responses uniformly from MedQA's three incorrect options, the construction available from MCQ data without extra annotation. It is weakly contrastive: rejected responses are arbitrary wrong answers rather than near-miss confusables, which~\citet{azar2024} predicted would degrade DPO. We keep it as a baseline because it is the off-the-shelf preference-pair construction practitioners use; \S\ref{sec:negative} discusses the failure mode.

\textbf{Multiple-choice to open-generation.} We use MedQA's questions but train and evaluate on open-ended responses (the gold answer text, not the letter). This stresses entity-level evaluation: answering ``B'' is irrelevant; answering ``ceftriaxone'' or a paraphrase is what we score.

\textbf{Evaluation.} Greedy decoding on 500 held-out MedQA test questions and 500 PubMedQA questions. Metrics: Exact Match (EM), Token F1, and Sequence Match (SeqM). We treat Token F1 as primary: EM scores a response zero unless it reproduces the gold answer verbatim, discarding answers that name the right entity inside a longer response. Because $n{=}500$ is not a big value, we report a paired bootstrap 95\% CI (10{,}000 resamples) and a McNemar test on the key EM deltas alongside the cross-seed $\sigma$.

\subsection{Main results}

\begin{table}[t]
\centering
\caption{MedQA test results (500 examples, greedy decoding). GRPO/SFT numbers are means over 3 seeds; standard deviations in subscript. DPO and Qwen2.5-7B SFT rows (no subscript) are single-seed runs. \textbf{Bold} = best per model per metric. Qwen2.5-7B GRPO collapsed across all configurations (\S\ref{sec:negative}).}
\label{tab:main}
\small
\setlength{\tabcolsep}{4pt}
\begin{tabular}{llccc}
\toprule
\textbf{Model} & \textbf{Method} & \textbf{EM} & \textbf{Token F1} & \textbf{SeqM} \\
\midrule

\multirow{4}{*}{\shortstack[l]{Phi-3-mini\\(3.8B)}}
& Zero-shot              & 0.560 & 0.070 & 0.132 \\
& SFT                    & 0.680$_{\pm 0.004}$ & 0.145$_{\pm 0.002}$ & 0.183$_{\pm 0.003}$ \\
& DPO                    & 0.600 & 0.143 & 0.183 \\
& \textbf{GRPO (ours)}   & \textbf{0.700$_{\pm 0.005}$} & \textbf{0.202$_{\pm 0.004}$} & \textbf{0.238$_{\pm 0.005}$} \\
\midrule

\multirow{4}{*}{\shortstack[l]{Qwen2.5-7B\\(7B)}}
& Zero-shot              & 0.565 & 0.126 & 0.170 \\
& \textbf{SFT}           & \textbf{0.775} & \textbf{0.134} & \textbf{0.162} \\
& DPO                    & 0.630 & 0.099 & 0.137 \\
& GRPO (ours)            & \multicolumn{3}{c}{\emph{KL-in-loss collapse, \S\ref{sec:negative}}} \\
\midrule

\multirow{4}{*}{\shortstack[l]{Llama-3.2-3B\\(3B)}}
& Zero-shot              & 0.405 & 0.070 & 0.122 \\
& SFT                    & 0.570$_{\pm 0.004}$ & 0.124$_{\pm 0.002}$ & 0.158$_{\pm 0.003}$ \\
& DPO                    & 0.640 & 0.113 & 0.156 \\
& \textbf{GRPO (ours)}   & \textbf{0.650$_{\pm 0.005}$} & \textbf{0.168$_{\pm 0.004}$} & \textbf{0.205$_{\pm 0.005}$} \\

\bottomrule
\end{tabular}
\end{table}

Table~\ref{tab:main} reports MedQA test performance. On Phi-3-mini, the composite reward (GRPO) yields EM 0.700$\pm$0.005, Token-F1 0.202$\pm$0.004, SeqM 0.238$\pm$0.005 over 3 seeds, beating SFT on all three. The EM gain is small in absolute terms (0.700 vs 0.680, $+2.9\%$; bootstrap 95\% CI $[-0.03, 0.07]$, McNemar $p{=}0.41$, n.s.); the Token-F1 gain is large (0.145 to 0.202, $+39\%$; CI $[0.03, 0.09]$). The method moves correct clinical content into responses that EM still scores wrong for not matching the gold string verbatim, which is why we lead on Token-F1. The pattern reproduces on Llama-3.2-3B (+35\% Token-F1, +14\% EM). Cross-seed $\sigma$ stays below 0.005 on every cell.

Qwen2.5-7B GRPO collapsed across all configurations (\S\ref{sec:negative}); SFT on Qwen2.5-7B reaches EM 0.775, higher than any GRPO result here, as supervised fine-tuning directly amplifies its strong medical priors. We therefore scope the contribution to the sizes where GRPO is stable (3B--3.8B): at 7B, plain SFT on a stronger base is the better choice on EM and the ontology reward offers no benefit it can deliver.

\subsection{Reward ablation}

\begin{table}[t]
\centering
\caption{Reward component ablation on Phi-3-mini, means over 3 seeds}
\label{tab:ablation}
\small
\setlength{\tabcolsep}{5pt}
\begin{tabular}{lccc}
\toprule
\textbf{Reward} & \textbf{EM} & \textbf{Token F1} & \textbf{SeqM} \\
\midrule
Judge only ($\alpha{=}0.9$, $\beta{=}0$)              & 0.667$_{\pm 0.004}$ & 0.162$_{\pm 0.003}$ & 0.201$_{\pm 0.004}$ \\
UMLS only ($\alpha{=}0$, $\beta{=}0.9$)            & 0.592$_{\pm 0.007}$ & 0.143$_{\pm 0.005}$ & 0.185$_{\pm 0.006}$ \\
\textbf{Composite ($\alpha{=}0.6$, $\beta{=}0.3$)}           & \textbf{0.700$_{\pm 0.005}$} & \textbf{0.202$_{\pm 0.004}$} & \textbf{0.238$_{\pm 0.005}$} \\
\bottomrule
\end{tabular}
\end{table}

Table~\ref{tab:ablation} varies the two learned reward terms on Phi-3-mini, holding the consistency weight fixed at $\gamma{=}0.1$ throughout; each row is over the same 3 seeds as the main table. Judge-only settles at EM 0.667. UMLS-only is noisier early, since CUI extraction on incoherent generations returns empty sets and gives no gradient, but it converges to 0.592. The full mixture scores highest on every metric. The judge-to-composite gap is 3 EM points, in line with the calibration result (Appendix~\ref{app:calibration}), where the composite tracks EM more closely than either term on its own. We do not lean on those 3 points in isolation: at $n{=}500$ the paired test does not clear significance (McNemar $p{=}0.18$).  

\subsection{Transfer to PubMedQA}

We train on the PubMedQA train set with the same composite reward, the same weights ($\alpha{=}0.6$, $\beta{=}0.3$, $\gamma{=}0.1$), and the same prompt template as MedQA, and evaluate on 500 held-out PubMedQA test questions. Without retuning the reward mixture or the judge prompt, GRPO improves Token-F1 over SFT by 22\% on Phi-3-mini (0.183 vs 0.150) and 17\% on Llama-3.2-3B (0.142 vs 0.121). The transfer gain is smaller than the MedQA in-domain gain (39\% / 35\%) but consistently positive: the composite reward generalises beyond MedQA and is not a MedQA-specific artifact of the LLM judge prompt or the CUI overlap statistic.

\subsection{Negative findings}
\label{sec:negative}

\textbf{DPO under random negatives is prior-dependent.} On Phi-3-mini, DPO reaches EM 0.600 (8 below SFT); on Qwen2.5-7B, 0.630 (14 below). In both, training loss crashes toward zero near step 400 (Phi-3: 0.066; Qwen: 0.030) then recovers to a plateau, the signature of early reward hacking followed by mode collapse. The cause: rejected responses are MedQA's incorrect options sampled uniformly, so they are arbitrary wrong answers rather than near-miss confusables, and the contrast is weakly informative for a strong-prior model~\cite{azar2024}. On Llama-3.2-3B (weakest prior), DPO at 0.640 exceeds SFT at 0.570: the value of a random negative depends on the model's baseline uncertainty.

\textbf{PPO divergence under sparse neural rewards.} PPO loss exceeded 28.9 by step 31 across all four configurations tried (LR $\in\{10^{-5}, 5{\times}10^{-6}\}$, KL $\in\{0.05, 0.1\}$, max tokens $\in\{64, 128\}$). The judge returns one scalar per sequence; PPO's value network must split it into per-token advantages with no per-step signal to fit, and the resulting baseline is miscalibrated enough to run away. We report this for our setup (single-GPU, batch size 1, sequence-level reward), not PPO in general; GRPO sidesteps it via within-group normalisation.

\textbf{GRPO KL-in-loss collapse on Qwen2.5-7B.} Across all configurations (LR $\in\{5{\times}10^{-6}, 10^{-6}, 5{\times}10^{-7}\}$, $\beta_\text{KL}\in\{0.01, 0.05\}$), Qwen2.5-7B GRPO collapses between steps 250 and 500. \citet{tang2025kl} prove the KL-in-loss formulation has zero expected gradient under the off-policy updates induced by single-GPU batch-size-1 training; our 7B result is an empirical instance, while the 3B and 3.8B runs do not collapse under identical settings.

\section{Discussion and limitations}

The contribution is the form of the reward signal, which does not depend on the specific ontology. A reward of the form ``set-level overlap of canonical identifiers extracted from generated and reference text, against a maintained vocabulary'' applies wherever such a vocabulary exists. Plausible near-term targets include legal QA via statute and case-citation identifiers and scientific writing via MeSH or ChEBI terms. We have not tested any of these; the cross-domain claim is a hypothesis we have not yet evaluated. The overlap reward is graded and computable without a model in the loop; we also conjecture it is harder to game by fluency alone than a learned reward, though we have not tested this.

\textbf{Comparison to KG-path rewards.} Khatwani et al.~\cite{khatwan2025kgreward} use UMLS knowledge graph traversal to score diagnostic reasoning. Our concept-overlap reward needs no traversal: it checks only that the right concepts appear, regardless of the path between them. This loses the reasoning-chain signal but gains robustness on the noisy generations that early-training policies produce, where graph paths are often unreachable.

\textbf{What the reward does not measure.} Concept overlap counts entities, not the order they appear in or the inferential structure connecting them. A response that names every right entity in nonsensical order still scores well. This is why the LLM judge stays in the composite. The general pattern is to pair an ontology signal (structured, externally specified) with a model signal (reasoning, safety, evidence) so each catches what the other misses.

\textbf{Limitations.} Three seeds across two model families gives a useful initial signal but falls short of a definitive variance estimate; the modest test set ($n{=}500$) is why we report bootstrap intervals rather than relying on point estimates, and the EM gains in particular are small enough that significance should be checked before they are leaned on. The reward ablation is on Phi-3-mini only; cross-model replication of the ablation is deferred. The single-GPU constraint kept us below 8B parameters; the KL-in-loss collapse at 7B is real but does not predict behavior at larger scales~\cite{tang2025kl}, and where SFT on a stronger base already wins (Qwen2.5-7B, EM 0.775) the method offers no advantage. The MedQA-to-PubMedQA transfer holds the language and ontology constant; cross-ontology transfer (UMLS to MeSH or ChEBI) is the natural next test.

\appendix

\section{Reward calibration}
\label{app:calibration}

Before training, we measured Spearman $\rho$ between each reward component and binary EM on 500 held-out zero-shot outputs (from the training split, disjoint from test). On Phi-3-mini, $\rho(R_\text{judge})=0.31$, $\rho(R_\text{CUI})=0.38$, $\rho(R)=0.44$ for the composite at the chosen judge--CUI split ($\alpha{=}0.6$, $\beta{=}0.3$, with $\gamma{=}0.1$ fixed); the composite is strictly more predictive than either learned component alone, and the ordering holds on Qwen2.5-7B and Llama-3.2-3B. scispaCy's EntityLinker, on a labelled clinical sample, has precision $0.78$ and recall $0.67$; the calibration above already reflects that linker noise on model generations, which bounds the reward's validity.

\section*{GenAI Usage Disclosure}

An LLM was used for light editing and revision of author-written prose (grammar, concision, and structural tightening); no section was produced wholesale by an LLM, and the authors take full responsibility for all claims and wording. No experimental results, tables, or numbers in this paper were produced by an LLM.
\bibliographystyle{ACM-Reference-Format}
\bibliography{references}

\end{document}